\documentclass{article}

\usepackage{arxiv}
\usepackage[utf8]{inputenc}
\usepackage[T1]{fontenc}
\usepackage{hyperref}
\usepackage{url}
\usepackage{booktabs}
\usepackage{amsfonts}
\usepackage{amsmath}
\usepackage{nicefrac}
\usepackage{microtype}
\usepackage{graphicx}
\usepackage{natbib}
\usepackage{tikz}
\usepackage{pgfplots}
\usepackage{placeins}
\pgfplotsset{compat=1.17}
\usetikzlibrary{arrows.meta}
\usepackage{caption}

\title{AraSSM: A Bidirectional State-Space Encoder for Arabic Masked Language Modeling}

\author{
	Ahmed Amine Aliane\thanks{Corresponding author: aaliane781@gmail.com} \\
	Arabic Institute for Translation, Algiers, Algeria \\
	CERIST, Algeria \\
	\And
    Hassina Aliane \\
	CERIST, Algeria \\
    \And 
	Nasredine Semmar \\
	CEA LIST, Paris-Saclay, France \\
	}

\date{July 17, 2026}

\renewcommand{\shorttitle}{AraSSM: A Bidirectional SSM Encoder for Arabic}

\hypersetup{
	pdftitle={AraSSM: A Bidirectional State-Space Encoder for Arabic Masked Language Modeling},
	pdfsubject={cs.CL, cs.LG},
	pdfauthor={Ahmed Amine Aliane, Hassina Aliane, Nasredine Semmar},
	pdfkeywords={Arabic NLP, State Space Models, Mamba, Masked Language Modeling, Pretrained Encoders},
}

\begin{document}
	\maketitle
	
	\begin{abstract}
		Pretrained Transformer encoders such as AraBERT, MARBERT, and CAMeLBERT have become the standard
		backbone for Arabic natural language understanding, but their self-attention mechanism scales
		quadratically with sequence length, which limits efficiency on long documents. Mamba, a selective
		state-space model (SSM), offers linear-time sequence modeling as a competitive alternative to
		attention, yet no dedicated bidirectional Mamba encoder pretrained specifically for Arabic
		currently exists. We introduce AraSSM, a bidirectional Mamba encoder pretrained via masked
		language modeling on a corpus combining Arabic Wikipedia and CulturaX text, trained end-to-end
		on four consumer-grade NVIDIA RTX 2080Ti GPUs (11GB) over approximately ten days. We evaluate
		AraSSM by fine-tuning on four established Arabic NLU benchmarks covering sentiment classification
		(HARD), named entity recognition (ANERcorp), extractive question answering (ARCD), and natural
		language inference (XNLI-ar), following the per-task evaluation protocol introduced by AraBERT,
		and report results as mean $\pm$ standard deviation across three fine-tuning seeds. AraSSM
		matches or exceeds published base-sized Transformer baselines on sentiment classification
		($96.37 \pm 0.03\%$ accuracy on HARD), is competitive on extractive QA ($32.19 \pm 1.07$ EM,
		$63.79 \pm 0.25$ F1 on ARCD) and named entity recognition ($81.54 \pm 0.30$ entity-level F1 on
		ANERcorp), and trails the base-sized Transformer range on natural language inference
		($72.83 \pm 0.07\%$ accuracy on XNLI-ar), despite being trained entirely from scratch on
		consumer hardware rather than large-scale accelerator clusters.
	\end{abstract}
	
	\keywords{Arabic NLP \and State Space Models \and Mamba \and Masked Language Modeling \and Pretrained Encoders}
	
	\section{Introduction}
	
	Pretrained Transformer encoders have become the dominant paradigm for Arabic natural language
	understanding. Models such as AraBERT \citep{antoun2020arabert}, ARBERT and MARBERT
	\citep{abdulmageed2021arbert}, and CAMeLBERT \citep{inoue2021interplay} have established strong
	baselines across a wide range of Arabic NLU tasks by adapting the BERT pretraining recipe
	\citep{devlin2019bert} to Arabic-specific tokenization and corpora. However, the self-attention
	mechanism at the core of these models scales quadratically with sequence length. This imposes a
	hard memory and compute ceiling on the context these encoders can process, and requires explicit
	truncation for documents beyond the model's maximum sequence length.
	
	Mamba \citep{gu2023mamba}, a selective state-space model (SSM), has recently emerged as a
	competitive alternative to attention-based architectures, offering linear-time sequence modeling
	with hardware-aware parallelism, and matching or exceeding Transformer performance across
	language, audio, and genomics modalities. Domain- or language-specific adaptations of Mamba have
	already demonstrated the value of pretraining an SSM backbone directly on specialized data, rather
	than relying on a general-purpose checkpoint. BioMamba \citep{biomamba2024} for biomedical text and
	ClinicalMamba \citep{yang2024clinicalmamba} for longitudinal clinical notes both show that a Mamba
	encoder pretrained on in-domain text outperforms both general-domain Mamba checkpoints and
	comparably sized Transformer baselines on their respective tasks. Despite this precedent, no
	equivalent effort currently exists for Arabic. Recent work applying Mamba/SSM architectures to
	Arabic text has done so without Arabic-specific pretraining, and reports that Mamba-based models
	lag behind smaller, Arabic-pretrained Transformer encoders as a direct consequence
	\citep{alzawqari2025neural}. This suggests that the architecture itself is not the limiting factor,
	and that the missing ingredient is a dedicated Arabic pretraining effort analogous to AraBERT's for
	the Transformer family.
	
	A second, largely orthogonal motivation for this work is compute accessibility. The Arabic
	Transformer encoders discussed above were pretrained on large-scale accelerator clusters. In this paper, we introduce
    AraSSM, a bidirectional Mamba encoder pretrained from scratch for Arabic via a masked language modeling (MLM) objective.
    AraSSM is pretrained and fine-tuned entirely on four consumer-grade RTX 2080Ti GPUs (11GB
	each, Turing architecture), a hardware budget within reach of a single academic lab rather than an
	industrial research group. We view the resulting comparison, a from-scratch SSM encoder trained on
	consumer hardware against Transformer encoders trained on considerably larger compute budgets, as
	informative in its own right, independent of whether AraSSM matches or trails those baselines on
	any individual task.
	
	Since Mamba is inherently causal, we adopt a
	bidirectional block design that runs a forward and a backward selective-scan mixer over each
	sequence and merges their outputs, following the general recipe used in bidirectional Mamba
	adaptations for other modalities, to obtain full bidirectional context while retaining $O(L)$
	complexity in sequence length $L$, instead of the $O(L^2)$ complexity of attention. AraSSM is
	pretrained on a corpus combining Arabic Wikipedia and CulturaX, cleaned and deduplicated at the
	document-chunk level. Our contributions are summarized as follows:
	
	\begin{enumerate}
		\item We introduce, to the best of our knowledge, the first bidirectional Mamba/SSM encoder
		pretrained specifically for Arabic via masked language modeling.
		\item We evaluate AraSSM on four established Arabic NLU benchmarks, covering sentiment
		classification (HARD), named entity recognition (ANERcorp), extractive question answering
		(ARCD), and natural language inference (XNLI-ar), under a fine-tuning protocol directly
		comparable to the one introduced by AraBERT and adopted by subsequent Arabic Transformer
		encoders \citep{antoun2020arabert, antoun2021araelectra}, reporting results across three
		fine-tuning seeds to quantify result stability.
		\item We situate these results against a compute budget of four consumer-grade RTX 2080Ti GPUs
		for both pretraining and fine-tuning, in contrast to the accelerator-cluster budgets typically
		used to pretrain Arabic Transformer encoders, and discuss the resulting accuracy/compute
		trade-off.
	\end{enumerate}
	
	\section{Related Work}
	
	Our work lies at the intersection of Arabic pretrained language models and state-space sequence
	modeling. We present in this section an overview of key advances in these two areas, as well as
	prior efforts to pretrain Mamba-based encoders for specific domains and languages.
	
	\subsection{Arabic Pretrained Language Models}
	
	Arabic-specific pretraining has followed the broader trajectory of Transformer-based language
	models. AraBERT \citep{antoun2020arabert} was among the first dedicated Arabic encoders, adapting
	the BERT pretraining recipe \citep{devlin2019bert} to a large Modern Standard Arabic (MSA) corpus
	with Arabic-specific preprocessing and tokenization. ARBERT and MARBERT
	\citep{abdulmageed2021arbert} extended this line of work, with MARBERT specifically targeting
	Arabic dialectal variation through large-scale pretraining on dialectal Twitter data, in contrast
	to AraBERT's MSA-focused corpus. CAMeLBERT \citep{inoue2021interplay} systematically studied the
	effect of pretraining data variety (MSA, dialectal, and mixed) on downstream task performance,
	while AraELECTRA \citep{antoun2021araelectra} adapted the more sample-efficient ELECTRA pretraining
	objective to Arabic. Benchmarking efforts such as ALUE \citep{seelawi2021alue} and ORCA
	\citep{elmadany2023orca} have since standardized evaluation across this growing family of Arabic
	encoders by consolidating dozens of individual datasets into composite leaderboards. In this work,
	we evaluate AraSSM on four of the individual, widely-used benchmarks that underlie these
	consolidated efforts (Section~4), rather than the composite ORCA/ALUE scores themselves, so that
	our results remain directly comparable, per task, to the original numbers reported by AraBERT and
	AraELECTRA. All of the encoders discussed above, however, share the same underlying Transformer
	architecture and its associated quadratic attention cost as well as the substantial compute budgets required for pretraining at scale.
	
	\subsection{State Space Models and Mamba}
	
	State space models (SSMs) offer an alternative to attention with linear-time complexity in
	sequence length. Structured State Space models (S4) \citep{gu2021s4} first demonstrated that SSMs
	could capture long-range dependencies with the parallelizability of Transformers while avoiding
	their quadratic cost. Mamba \citep{gu2023mamba} introduced a selective state-space mechanism with
	input-dependent parameterization, allowing the model to selectively propagate or forget
	information along the sequence, and combined this with a hardware-aware parallel scan
	implementation, closing much of the performance gap with attention-based models while retaining
	$O(L)$ complexity. Because Mamba's recurrence is inherently causal, adapting it to encoder-style
	bidirectional tasks requires an explicit architectural modification. Prior work in other modalities
	has addressed this by running paired forward and backward mixers and merging their outputs, an
	approach we adopt for AraSSM's encoder blocks (Section~3).
    This bidirectional adaptation has been explored in vision (Vim \citep{zhu2024visionmamba}, VMamba \citep{liu2024vmamba}) and biomedical NLP (BioMamba \citep{biomamba2024}), demonstrating that the approach generalizes across modalities
    while preserving Mamba's linear-time complexity.
	
	\subsection{Domain- and Language-Specific Mamba Pretraining}
	
	A growing body of work has shown that pretraining Mamba directly on in-domain text, rather than
	fine-tuning a general-purpose checkpoint, yields better downstream performance for specialized
	domains. BioMamba \citep{biomamba2024} pretrains a Mamba-based encoder on biomedical literature,
	while ClinicalMamba \citep{yang2024clinicalmamba} pretrains on longitudinal clinical notes, in both
	cases outperforming both general-domain Mamba and comparably sized Transformer baselines on
	in-domain tasks. To the best of our knowledge, no equivalent language-specific pretraining effort
	exists for Arabic. The closest related work applies Mamba/SSM architectures to an Arabic task
	(argument classification in competitive debates) without Arabic-specific pretraining, and finds
	that, absent such pretraining, Mamba-based models underperform smaller, dedicated Arabic
	Transformer encoders \citep{alzawqari2025neural}, a finding consistent with the domain-pretraining
	literature above, and the central gap this paper addresses.
	
	\section{Methodology}
	
	\subsection{Model Architecture}
	
\begin{figure}[t]
		\centering
		\begin{tikzpicture}[
			box/.style={draw, rounded corners, minimum width=2.6cm, minimum height=0.8cm, align=center},
			arrow/.style={-{Latex}}
			]
			\node[box] (input) at (0,0) {Input $h$};
			\node[box] (norm) at (0,-1.2) {LayerNorm};
			\node[box] (fwd) at (-1.8,-2.6) {Forward\\SSM};
			\node[box] (bwd) at (1.8,-2.6) {Backward\\SSM (flipped)};
			\node[box] (merge) at (0,-4) {Concat + $W_{\text{merge}}$};
			\node[box] (res1) at (0,-5.2) {$+$ Residual};
			\node[box] (ffn) at (0,-6.4) {FFN (LayerNorm $\to$ GELU MLP)};
			\node[box] (res2) at (0,-7.6) {$+$ Residual};
			\node[box] (output) at (0,-8.8) {Output $h'$};
			
			\draw[arrow] (input) -- (norm);
			\draw[arrow] (norm) -- (fwd);
			\draw[arrow] (norm) -- (bwd);
			\draw[arrow] (fwd) -- (merge);
			\draw[arrow] (bwd) -- (merge);
			\draw[arrow] (merge) -- (res1);
			\draw[arrow] (res1) -- (ffn);
			\draw[arrow] (ffn) -- (res2);
			\draw[arrow] (res2) -- (output);
			\draw[arrow] (input.east) to[out=0,in=90] (res1.east);
			\draw[arrow] (res1.east) to[out=0,in=90] (res2.east);
		\end{tikzpicture}
		\caption{One AraSSM encoder block. Forward and backward selective-scan mixers run over the same
			normalized input; their outputs are concatenated and merged before the residual and feed-forward
			sublayers.}
		\label{fig:bimamba-block}
	\end{figure}
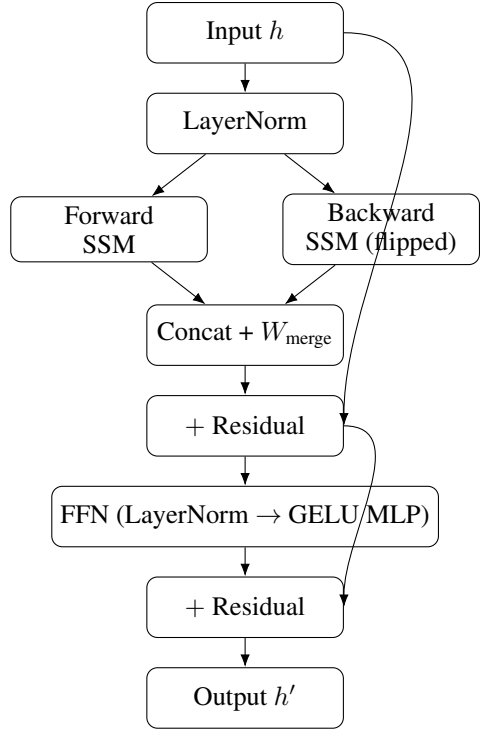

	AraSSM follows the bidirectional block design outlined in Section~2.2: each layer runs a forward
	selective-scan mixer and a backward selective-scan mixer over the same normalized hidden state,
	then merges the two outputs through a learned linear projection before passing the result to a
	position-wise feed-forward sublayer. Figure~\ref{fig:bimamba-block} illustrates the full block.
	
	Formally, the selective-scan mixer maps an input sequence $x_{1:L}$ to an output sequence
	$y_{1:L}$ through a discretized, input-dependent linear state-space recurrence:
	\begin{align}
		h_t &= \bar{A}_t h_{t-1} + \bar{B}_t x_t, \label{eq:ssm-state} \\
		y_t &= C_t h_t, \label{eq:ssm-output}
	\end{align}
	where $h_t \in \mathbb{R}^{d_{\text{state}}}$ is the hidden state at position $t$, and
	$\bar{A}_t$, $\bar{B}_t$, $C_t$ are obtained by discretizing continuous parameters $A$, $B$, $C$
	with a step size $\Delta_t$ that is itself a function of $x_t$, making the recurrence selective:
	\begin{equation}
		\Delta_t, B_t, C_t = f_\Delta(x_t), \, f_B(x_t), \, f_C(x_t),
	\end{equation}
	so the model can choose, per token, how much of the running state to retain versus overwrite.
	Given hidden state $h \in \mathbb{R}^{L \times d}$ at a block's input, the bidirectional block
	computes
	\begin{equation}
		h_{\text{fwd}} = \text{SSM}(h), \qquad h_{\text{bwd}} = \text{flip}\big(\text{SSM}(\text{flip}(h))\big),
	\end{equation}
	and merges the two directions through a learned projection $W_{\text{merge}} \in \mathbb{R}^{2d
		\times d}$:
	\begin{equation}
		h' = h + \text{Dropout}\big(W_{\text{merge}} \, [h_{\text{fwd}} \, ; \, h_{\text{bwd}}]\big).
	\end{equation}
	This mirrors the general bidirectional Mamba recipe adopted in other modalities, and lets each
	token attend to context on both sides while every mixer call remains a causal, linear-time scan.
	
	For the padded positions introduced by batching variable-length sequences, we apply the attention
	mask twice inside each block: once before the backward mixer, so that padding tokens at the end of
	a sequence do not leak non-zero values into the reversed scan, and once after the block's output,
	so that padding contributes nothing to the residual stream at any depth. Without this masking,
	backward mixers would silently condition real tokens on the identity of arbitrary padding content,
	making the model's behavior at real positions depend on batch composition rather than only on the
	sequence itself. Concretely, rather than a naive reversal of the padded sequence, only the valid
	(non-padding) positions of each example are reversed in place, leaving padding at the tail of the
	sequence exactly where it already sits; this reduces to an ordinary full-sequence flip whenever a
	batch contains no padding, which is always the case during pretraining, where documents are
	packed into fixed-length, unpadded chunks (Section~3.2), so this fine-tuning-time masking detail
	does not alter or invalidate the pretrained checkpoint used as the starting point for Section~4.

    \begin{table}[h]
    	\centering
    	\begin{tabular}{lcccc}
    		\toprule
    		Model & Layers & Hidden size & Params & Backbone \\
    		\midrule
    		AraBERT-base & 12 & 768 & $\sim$110M & Transformer \\
    		AraSSM & 12 & 512 & 105.0M & SSM (Mamba) \\
    		\bottomrule
    	\end{tabular}
    	\caption{AraSSM and AraBERT-base are close in overall size despite AraSSM's smaller hidden
		dimension, since each AraSSM layer runs two selective-scan mixers (forward and backward)
		rather than one attention block.}
    	\label{tab:model-size-comparison}
      \end{table}

	AraSSM uses a hidden size of 512, 12 layers, a state dimension of 16, a convolutional kernel width
	of 4, and an expansion factor of 2 inside each Mamba mixer, matching the depth and width of
	BERT-base and, by extension, AraBERT-base \citep{antoun2020arabert}, so that AraSSM is directly
	comparable to our Transformer baselines in nominal parameter count and layer depth. We choose a hidden size of 512, slightly 
    smaller than AraBERT's 768, to fit the full bidirectional block within the 11GB VRAM budget while maintaining 12 layers. 
    Table~\ref{tab:hyperparams} summarizes the full model and pretraining configuration. This depth-width choice warrants some
	justification, since it runs against a widely noted rule of thumb for causal Mamba language
	models: the original Mamba implementation reports that Mamba's layer count typically needs to
	double that of a similarly sized Transformer, because two Mamba blocks are needed to match the
	sequence-mixing plus channel-mixing capacity of one Transformer layer, which combines a multi-head
	attention block and an MLP block \citep{gu2023mamba}. Taken at face value, this would suggest a
	12-layer Mamba encoder is undersized relative to a 12-layer Transformer. That guidance, however, is
	derived from unidirectional, causal Mamba stacks. Each AraSSM block instead runs two full selective
	scans per layer, one forward and one backward, rather than the single causal scan the
	depth-doubling rule of thumb assumes, so a 12-layer bidirectional AraSSM stack already performs as
	many sequence scans as a 24-layer causal Mamba stack would. We treat this as a design rationale for
	keeping AraSSM's depth matched to AraBERT-base rather than doubled, not as an established
	equivalence, and we report AraSSM's downstream results against these matched-configuration
	baselines in Section~4 as the direct empirical test of it.
	
	\begin{table}[h]
    	\centering
    	\begin{tabular}{lc}
    		\toprule
    		Parameter & Value \\
    		\midrule
    		Layers & 12 \\
    		Hidden size ($d_{\text{model}}$) & 512 \\
    		State dimension ($d_{\text{state}}$) & 16 \\
    		Conv kernel width ($d_{\text{conv}}$) & 4 \\
    		Expansion factor & 2 \\
    		Vocabulary size & 64,000 \\
    		Max sequence length (training) & 512 \\
    		Dropout & 0.1 \\
    		Pretraining corpus size & $\sim$80GB (79.6GB train / 0.4GB val) \\
    		Optimizer & AdamW \\
    		Learning rate & 3e-4 \\
    		Weight decay & 0.01 \\
    		Warmup steps & 10,000 \\
    		Training steps & 200,000 \\
    		Batch size (per GPU) & 8 \\
    		Gradient accumulation & 8 \\
    		Effective batch size & 256 \\
    		Precision & fp16 \\
    		\bottomrule
    	\end{tabular}
    	\caption{AraSSM model and pretraining configuration, from \texttt{base.yaml}.}
    	\label{tab:hyperparams}
    \end{table}
	
	The token embedding matrix is tied to the output projection of the MLM head, and weights are
	initialized with a truncated-normal scheme (std $=$ 0.02), rather than PyTorch's default
	initialization, since the default leaves the 64k-way output distribution badly overconfident at
	step zero and produces an initial loss well above the $\ln(|V|)$ expected from a uniform
	prediction, which in turn destabilizes early optimization. Gradient checkpointing is enabled at the
	block level during training, which we found necessary to keep peak memory within the budget of the
	11GB cards used in our infrastructure (Section~3.4) at the batch sizes we target.
	
	\subsection{Pretraining Corpus}
	
	AraSSM is pretrained on a combination of Arabic Wikipedia and the Arabic portion of CulturaX \citep{nguyen2023culturax}, an
	already-deduplicated web corpus derived from mC4 and OSCAR. We stream both sources rather than
	downloading them in full, avoiding provisioning disk for the raw, undeduplicated data and letting
	us stop consumption once a target size has been reached. Wikipedia is consumed first for its
	cleaner, encyclopedic register, and CulturaX is drawn on afterward to reach the full target corpus
	size, so that the same total-token budget contains a larger fraction of high-quality text than a
	purely web-sourced corpus of the same size would.
	
	Each document passes through a lightweight cleaning stage before being added to the corpus: Arabic
	diacritics are stripped, URLs are removed, whitespace is normalized, and documents are discarded
	if they fall under a minimum length of 20 words or if fewer than 70\% of their characters fall in
	the Arabic Unicode block. Documents that survive cleaning are then split into chunks of at most
	400 words before being written out, rather than kept as single, whole-document lines. This step
	matters because the training-side tokenizer truncates each line independently to the model's
	maximum sequence length: without chunking, long Wikipedia articles and CulturaX web pages, which
	routinely run to several hundred or several thousand words, would either be silently truncated
	past their first few hundred tokens, discarding most of the text a document contributed to the
	corpus, or would push the great majority of batches to the maximum sequence length, inflating and
	destabilizing GPU memory use. The 400-word threshold is chosen as an approximate proxy for the
	model's 512-token maximum sequence length, given that the AraBERT-derived subword tokenizer used
	downstream (Section~3.3) produces roughly one to 1.3 tokens per word depending on register,
	leaving headroom for special tokens and subword splitting. This threshold is a heuristic specific
	to our pipeline rather than a value drawn from prior work, since it depends on the token-to-word
	ratio of the particular tokenizer and register mix used here.
	
	Deduplication is applied at the chunk level using a bounded-memory Bloom filter, so that repeated
	boilerplate spanning multiple documents, such as navigation text or template content common on web
	pages, is caught even when the surrounding document differs. The resulting corpus is written to
	sharded text files rather than a single file, so that downstream training code never has to hold
	the full corpus in memory, and a small fraction of chunks is routed to a held-out validation split
	rather than the training split. The pipeline itself is designed to scale to a stated disk
	budget rather than a fixed token count, stopping once a configurable target size of cleaned
	text has been written; the resulting corpus totals approximately 80GB (79.6GB train, 0.4GB
	validation), as summarized in Table~\ref{tab:hyperparams} and Section~3.4.
	
	\subsection{Pretraining Objective and Tokenization}
	
	AraSSM is pretrained with the standard masked language modeling (MLM) objective introduced by BERT
	\citep{devlin2019bert}. For each training example, 15\% of tokens are selected for prediction; of
	these, 80\% are replaced with a mask token, 10\% are replaced with a random token drawn uniformly
	from the vocabulary, and the remaining 10\% are left unchanged, following the original BERT
	masking scheme \citep{devlin2019bert}. The model is trained to predict the original identity of
	each selected token from the corrupted sequence, and the cross-entropy loss is computed only over
	the selected positions, with all other positions excluded via an ignore index. Padding positions
	are always excluded from masking, so that the model is never asked to predict a pad token.
	
	Given a corrupted sequence $\tilde{x}_{1:L}$ and the set of masked positions $\mathcal{M}
	\subseteq \{1, \dots, L\}$, the model is trained to minimize
	\begin{equation}
		\mathcal{L}_{\text{MLM}} = -\frac{1}{|\mathcal{M}|} \sum_{t \in \mathcal{M}} \log
		p_\theta(x_t \mid \tilde{x}_{1:L}),
	\end{equation}
	where $p_\theta$ is the model's predicted distribution over the vocabulary at position $t$,
	obtained from the tied output projection described above.
	
	Tokenization uses the pretrained AraBERTv02 tokenizer (\texttt{aubmindlab/bert-base-arabertv02})
	from \citet{antoun2020arabert}, giving a vocabulary of 64,000 subword units. We adopt this existing
	tokenizer rather than training a new one from scratch, both to keep the corpus preprocessing
	pipeline decoupled from tokenizer choice and to keep results comparable to AraBERT-family baselines
	that use the same vocabulary. Sequences are truncated to a maximum length of 512 tokens at the
	point of tokenization, consistent with the model's maximum position length.
	
	\subsection{Optimization and Distributed Training}

    We use AdamW with a learning rate of 3e-4, weight decay of 0.01, and a linear warmup of 10,000
    steps followed by linear decay to zero over the remainder of training, following the general
    optimization recipe used for BERT-style MLM pretraining \citep{devlin2019bert}. Both the batch
    size and the numerical precision used here are chosen to fit the model within the 11GB memory
    budget of our target hardware, four NVIDIA RTX 2080Ti GPUs (Turing architecture). At sequence
    length 512, a 12-layer, 512-dimensional bidirectional Mamba stack leaves little headroom on an
    11GB card once activations, gradients, and optimizer state are accounted for, so we use a small
    per-GPU batch size of 8 and recover a larger effective batch through 8 steps of gradient
    accumulation, for a global effective batch of 256 sequences per optimizer step across the 4 GPUs.
    Mixed-precision training is performed in fp16, the only tensor-core-accelerated mixed-precision
    mode available on Turing-architecture GPUs. Gradient scaling is handled through PyTorch's
    \texttt{GradScaler} to guard against fp16 underflow, and gradients are clipped to a maximum norm
    of 1.0 before each optimizer step.
    
    Because the 2080Ti's 11GB memory budget is small relative to the model and batch configuration
    described above, several choices in Section~3.1 and this section are made specifically to fit
    training within that budget: gradient checkpointing at every block, fp16 mixed precision, and a
    modest per-GPU batch size compensated for by gradient accumulation rather than a larger
    single-step batch. Checkpointing saves the full model, optimizer, scaler, and scheduler state
    every 5,000 optimizer steps, and training resumes automatically from the latest checkpoint if one
    is present, which we relied on given the intermittent availability of our compute infrastructure.
    Held-out validation loss, masked-token accuracy, and perplexity are computed on the main process
    at every checkpoint interval, over a fixed number of validation batches, without requiring the
    other ranks to participate, since evaluation involves only forward passes and no gradient
    synchronization.
    
    Before committing to the full 200,000-step run, we validated the training pipeline at a minimal
    scale to confirm the code path executes correctly end to end, then ran a short pilot using the
    full base architecture (2,000 steps, a slightly higher learning rate of 5e-4, and a shorter
    warmup) to stress-test real VRAM and throughput behavior at the target batch size before
    committing compute to the full schedule. Full pretraining ran for approximately ten days
    across the four GPUs described above, corresponding to roughly 960 GPU-hours
    ($4 \times 24 \times 10$), to process the $\sim$80GB pretraining corpus described in Section~3.2
    (79.6GB train, 0.4GB validation), a compute budget several orders of magnitude smaller than the
    accelerator-cluster budgets typically reported for training Arabic Transformer encoders of
    comparable size.
	
	\section{Experiments and Discussion}
	
	\subsection{Fine-Tuning Protocol}
	
	For every downstream task, we attach a thin task-specific head (mean-pooled linear classifier for
	sequence classification, per-token linear classifier for token classification, and a two-way
	linear span head for extractive QA) on top of the pretrained AraSSM encoder, and fine-tune the
	full model end-to-end with AdamW. Model selection follows standard practice: after each training
	epoch we evaluate on a held-out development split, and the checkpoint with the best development
	score is restored before computing the final metrics on the (never-touched-during-development)
	test split; all results reported below are therefore test-set scores of the best-on-dev
	checkpoint, not the final-epoch checkpoint. For extractive QA specifically, the predicted answer
	span at evaluation time is obtained via a joint start/end decoding pass restricted to the context
	portion of the input (excluding the question tokens), searching over the top-$k$ start and end
	logit positions and selecting the highest-scoring valid span up to a maximum answer length of 30
	tokens, following standard SQuAD-style span decoding. For every task, we fine-tune with three
	fixed seeds (42, 1337, 2024) and report the mean and standard deviation of the test
	metric across seeds, rather than a single run, since single-seed results can be misleadingly
	optimistic or pessimistic on smaller test sets (Section~4.6).

    \begin{table}[h]
    	\centering
    	\begin{tabular}{lccccc}
    		\toprule
    		Task & Batch size & LR & Epochs & Warmup steps & Max seq.\ length \\
    		\midrule
    		HARD & 16 & 2e-5 & 4 & -- & 256 \\
    		XNLI-ar & 16 & 2e-5 & 3 & -- & 128 \\
    		ANERcorp & 4 & 2e-5 & 5 & 50 & 128 \\
    		ARCD & 8 & 3e-5 & 10 & 50 & 384 \\
    		\bottomrule
    	\end{tabular}
    	\caption{Fine-tuning hyperparameters per downstream task. All tasks use AdamW with weight
    		decay 0.01, fp16 mixed precision, and the AraBERTv02 tokenizer; task-specific values above
    		were selected via light manual tuning per task rather than a shared search. Evaluation
    		batch sizes are 32 (HARD, XNLI-ar, ANERcorp) and 16 (ARCD).}
    	\label{tab:finetune-hyperparams}
    \end{table}

	\subsection{Sentiment Classification: HARD}
	
	The Hotel Arabic-Reviews Dataset (HARD) \citep{elnagar2018hard} consists of Arabic hotel reviews
	with 1--5 star ratings. Following \citet{antoun2020arabert}, we binarize the task by mapping
	ratings 1--2 to negative, ratings 4--5 to positive, and discarding neutral (rating 3) reviews. We
	fine-tune AraSSM directly from the pretrained MLM checkpoint, with no intermediate prefinetuning
	stage. Table~\ref{tab:hard-results} reports our results against comparable base-sized Transformer
	baselines reported by \citet{antoun2020arabert} on the same binarization scheme.
	
	\begin{table}[h]
		\centering
		\begin{tabular}{lcc}
			\toprule
			Model & Accuracy (\%) & Macro-F1 (\%) \\
			\midrule
			mBERT \citep{antoun2020arabert} & 95.0 & -- \\
			AraBERTv0.1/v1 \citep{antoun2020arabert} & 96.0 & -- \\
			AraSSM (ours, 3 seeds) & $96.37 \pm 0.03$ & $96.37 \pm 0.03$ \\
			\bottomrule
		\end{tabular}
		\caption{HARD (binary sentiment) results. Our split is a fixed, seeded 80/10/10 split (Section~4.1)
			and is not the original curated benchmark split used by \citet{antoun2020arabert}; results are
			informally, not strictly, comparable. AraSSM standard deviation is computed across seeds
			42, 1337, 2024.}
		\label{tab:hard-results}
	\end{table}

	\subsection{Named Entity Recognition: ANERcorp}
	\label{sec:anercorp}

    ANERcorp \citep{benajiba2007anercorp} is a token-level Arabic NER corpus annotated with the
    standard PER/LOC/ORG/MISC entity classes in BIO format. Since the corpus is distributed as a flat
    stream of (word, tag) pairs with no sentence boundaries, we reconstruct pseudo-sentences by
    splitting after sentence-final punctuation tagged \texttt{O}, with a hard cap of 120 words per run
    as a safety fallback, triggered in only 0.64\% of reconstructed pseudo-sentences (34/5{,}278 across
    train and test). We report entity-level precision, recall, and F1 computed with \texttt{seqeval},
    the standard entity-level metric used in the Arabic NER literature, rather than token-level
    accuracy. AraSSM is fine-tuned with a plain linear token-classification head, initialized from a
    checkpoint that was itself first fine-tuned on Polyglot-NER's Arabic silver-standard NER data
    \citep{alrfou2015polyglotner} before being fine-tuned on ANERcorp, following the standard
    silver-then-gold warm-start recipe used across the Arabic NER literature to compensate for
    ANERcorp's comparatively small size.
	
	\begin{table}[h]
		\centering
		\begin{tabular}{lccc}
			\toprule
			Model & Precision (\%) & Recall (\%) & F1 (\%) \\
			\midrule
			Base-sized Transformer average\textsuperscript{*} & -- & -- & 82--83 \\
			AraSSM (ours, 3 seeds) & $82.76 \pm 0.84$ & $80.38 \pm 0.64$ & $81.54 \pm 0.30$ \\
			\bottomrule
		\end{tabular}
		\caption{ANERcorp (4-class NER) results, plain linear classification head, full tagset including
			MISC. \textsuperscript{*}Average of AraBERTv2, ARBERT, and AraELECTRA with a plain
			classification head, as reported in a relabeling study of the dataset \citep{alduwais2024clean}.
			AraSSM precision consistently exceeds recall by 2--3 points across all three seeds,
			suggesting the model is somewhat conservative about proposing entity spans rather than
			over-predicting them.}
		\label{tab:anercorp-results}
	\end{table}
	
	\subsection{Extractive Question Answering: ARCD}
	\label{sec:arcd}
	
	The Arabic Reading Comprehension Dataset (ARCD) \citep{mozannar2019arcd} contains around 1,400
	crowd-sourced question-answer pairs over Arabic Wikipedia paragraphs. Given ARCD's small training
	set (roughly 700 examples after our fixed train/dev split), we follow a sequential warm-start
	recipe in the spirit of the one introduced alongside the dataset \citep{mozannar2019arcd}: rather
	than fine-tuning on Arabic-SQuAD alone, we first jointly fine-tune AraSSM on a combination of
	Arabic-SQuAD \citep{mozannar2019arcd} ($\sim$48K machine-translated QA pairs) and the Arabic subset
	of TyDi QA's Gold Passage secondary task \citep{clark2020tydiqa} (human-written questions over
	native, non-translated Arabic Wikipedia passages), motivated by machine translation introducing
	systematic ``translationese'' artifacts that a native-Arabic QA source can help offset; the
	resulting checkpoint is then fine-tuned on ARCD itself. This two-stage recipe is necessary because
	ARCD's training set alone is too small to train a randomly-initialized span-extraction head from
	scratch. Table~\ref{tab:arcd-results} reports results against representative base-sized Transformer
	baselines from \citet{antoun2021araelectra}, using the same EM/F1 metrics.
	
	\begin{table}[h]
		\centering
		\begin{tabular}{lcc}
			\toprule
			Model & EM & F1 \\
			\midrule
			Arabic-BERT-base \citep{antoun2021araelectra} & 30.5 & 62.2 \\
			ARBERT \citep{antoun2021araelectra} & 31.6 & 65.9 \\
			AraBERTv0.1 \citep{antoun2021araelectra} & 31.6 & 67.5 \\
			AraBERTv0.2-base \citep{antoun2021araelectra} & 32.8 & 66.5 \\
			AraSSM (ours, 3 seeds) & $32.19 \pm 1.07$ & $63.79 \pm 0.25$ \\
			AraBERTv0.2-large \citep{antoun2021araelectra} & 36.9 & 71.3 \\
			\bottomrule
		\end{tabular}
		\caption{ARCD (extractive QA) results. AraSSM is fine-tuned via the sequential
			Arabic-SQuAD+TyDiQA-ar $\to$ ARCD recipe described above; all baselines are base-sized
			models except AraBERTv0.2-large, included for reference only. Note that a single-seed run
			(seed 42) reached EM 35.0, above the 3-seed mean of 32.19; see Section~\ref{sec:discussion}
			for discussion of why the multi-seed mean, not the best single run, is the number we treat
			as representative.}
		\label{tab:arcd-results}
	\end{table}
	
	\subsection{Natural Language Inference: XNLI-ar}
	
	We additionally evaluate on the Arabic split of XNLI \citep{conneau2018xnli}, a 3-way natural
	language inference task (entailment / neutral / contradiction) over machine-translated
	premise-hypothesis pairs. AraSSM is fine-tuned directly from the pretrained MLM checkpoint with a
	mean-pooled linear classification head, packing the premise and hypothesis as a single sequence
	via the tokenizer's pair-encoding mode.
	
	\begin{table}[h]
		\centering
		\begin{tabular}{lcc}
			\toprule
			Model & Accuracy (\%) & Macro-F1 (\%) \\
			\midrule
			Base-sized Transformer range\textsuperscript{*} & 73--78 & -- \\
			AraSSM (ours, 3 seeds) & $72.83 \pm 0.07$ & $72.87 \pm 0.08$ \\
			\bottomrule
		\end{tabular}
		\caption{XNLI-ar (3-way NLI) results. \textsuperscript{*}Approximate range spanning several
			base-sized Arabic/multilingual Transformer encoders under full fine-tuning.
		}
		\label{tab:xnli-results}
	\end{table}

    \subsection{Result Summary and Baseline Comparison}
     
\begin{figure}[h]
    	\centering
    	\begin{tikzpicture}
    		\begin{axis}[
    			ybar,
    			bar width=22pt,
    			width=0.95\textwidth,
    			height=6cm,
    			symbolic x coords={HARD (Acc.), ANERcorp (F1), ARCD (EM), ARCD (F1), XNLI-ar (Acc.)},
    			xtick=data,
    			x tick label style={rotate=20, anchor=east, font=\small},
    			ylabel={AraSSM $-$ baseline (points)},
    			ymin=-3.5, ymax=1.5,
    			axis x line=middle,
    			axis y line=left,
    			enlarge x limits=0.12,
    			nodes near coords,
    			every node near coord/.append style={font=\scriptsize},
    			]
    			\addplot[fill=blue!60] coordinates {
    				(HARD (Acc.), 0.37)
    				(ANERcorp (F1), -0.96)
    				(ARCD (EM), 0.57)
    				(ARCD (F1), -1.74)
    				(XNLI-ar (Acc.), -2.67)
    			};
    		\end{axis}
    	\end{tikzpicture}
    	\caption{AraSSM's mean test score minus a representative base-sized Transformer reference point,
    		per task, in percentage points (or EM/F1 points for ARCD). Positive bars indicate AraSSM
    		exceeds the reference; negative bars indicate it trails. See Section~4.8 for the specific
    		reference values used per task.}
    	\label{fig:baseline-delta}
    \end{figure}
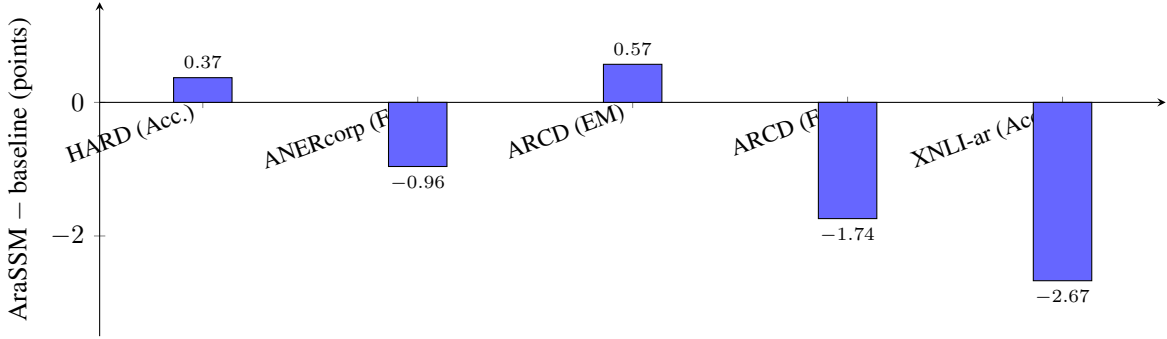

    Figure~\ref{fig:baseline-delta} summarizes AraSSM's position relative to base-sized Transformer
    baselines across all four benchmarks, expressed as the difference in percentage points between
    AraSSM's mean score and a representative baseline reference point per task (AraBERTv0.1 for HARD
    and ARCD-EM, the base-sized Transformer average for ANERcorp and ARCD-F1, and the midpoint of the
    73--78\% range for XNLI-ar).
     
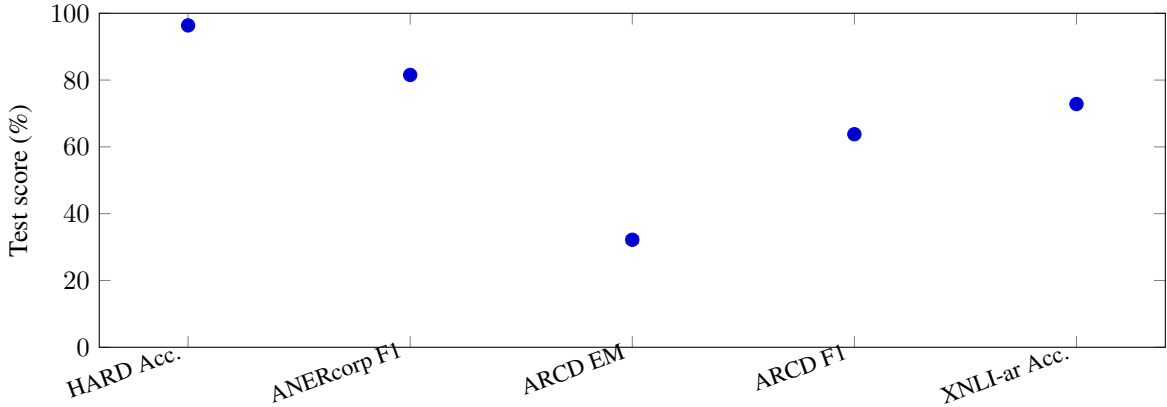
\begin{figure}[h]
    	\centering
    	\begin{tikzpicture}
    		\begin{axis}[
    			width=0.95\textwidth,
    			height=6cm,
    			symbolic x coords={HARD Acc., ANERcorp F1, ARCD EM, ARCD F1, XNLI-ar Acc.},
    			xtick=data,
    			x tick label style={rotate=20, anchor=east, font=\small},
    			ylabel={Test score (\%)},
    			ymin=0, ymax=100,
    			error bars/y dir=both,
    			error bars/y explicit,
    			only marks,
    			mark=*,
    			mark size=2.5pt,
    			]
    			\addplot+[error bars/.cd, y dir=both, y explicit]
    			coordinates {
    				(HARD Acc., 96.37) +- (0, 0.03)
    				(ANERcorp F1, 81.54) +- (0, 0.30)
    				(ARCD EM, 32.19) +- (0, 1.07)
    				(ARCD F1, 63.79) +- (0, 0.25)
    				(XNLI-ar Acc., 72.83) +- (0, 0.07)
    			};
    		\end{axis}
    	\end{tikzpicture}
    	\caption{AraSSM test-set score (mean $\pm$ standard deviation across seeds 42, 1337,
    		2024) for the primary metric of each task.}
    	\label{fig:seed-variance}
    \end{figure}

    Figure~\ref{fig:seed-variance} shows the mean test score $\pm 1$ standard deviation across the
    three fine-tuning seeds for each task's primary metric, illustrating that seed-to-seed variance is
    small and roughly proportional to test-set size: HARD and XNLI-ar, with the largest test sets,
    show negligible variance, while ARCD's exact-match metric, evaluated on a test set of roughly 350
    examples, shows the largest spread of the four.
     
    Taken together, these results establish that AraSSM achieves near-parity with base-sized
    Transformers on three of four benchmarks, with XNLI-ar as the notable outlier and AraSSM's
    largest gap relative to Transformer baselines. The next section examines how much of this
    pattern is attributable to pretraining itself; Section~4.8 then discusses the remaining open
    questions, including XNLI's machine-translated nature and the demands of sentence-pair semantic
    reasoning.

	\subsection{Ablation: Effect of Pretraining}
    To isolate the contribution of MLM pretraining itself, we fine-tune AraSSM on all four
    benchmarks starting from a randomly-initialized encoder, i.e. with the \texttt{pretrained\_ckpt}
    loading step skipped entirely, otherwise following the exact same fine-tuning protocol, task
    head, and three-seed evaluation described in Section~4.1. Table~\ref{tab:ablation-pretrain}
    reports the resulting test-set primary metric for each task, alongside the corresponding
    pretrained-encoder result from Sections~4.2--4.5, and the gap between the two.
    
    \begin{table}[h]
    	\centering
    	\begin{tabular}{lccc}
    		\toprule
    		Task (metric) & Pretrained & No pretraining & $\Delta$ \\
    		\midrule
    		HARD (Acc.\ \%) & $96.37 \pm 0.03$ & $95.31 \pm 0.09$ & $+1.06$ \\
    		XNLI-ar (Acc.\ \%) & $72.83 \pm 0.07$ & $52.79 \pm 0.32$ & $+20.04$ \\
    		ANERcorp (F1 \%) & $81.54 \pm 0.30$ & $48.03 \pm 0.20$ & $+33.51$ \\
    		ARCD (F1 \%) & $63.79 \pm 0.25$ & $12.51 \pm 0.05$ & $+51.28$ \\
    		\bottomrule
    	\end{tabular}
    	\caption{Effect of MLM pretraining, isolated by fine-tuning from a randomly-initialized
    		encoder instead of the pretrained AraSSM checkpoint. Mean $\pm$ standard
    		deviation across the same three seeds (42, 1337, 2024) used throughout Section~4.
    		$\Delta$ is pretrained minus no-pretraining.}
    	\label{tab:ablation-pretrain}
    \end{table}
    
    The size of this gap varies sharply across tasks, and the pattern is informative in its own
    right. On HARD, a large ($\sim$94K-example) sentiment classification task, the no-pretraining
    encoder reaches $95.31 \pm 0.09\%$ accuracy, only 1.06 points below the pretrained result: with
    enough labeled fine-tuning data, a randomly-initialized encoder can recover most of a sentiment
    classifier's signal from lexical cues alone. The gap widens substantially on XNLI-ar
    ($+20.04$ points), a task that requires modeling a semantic relation between two sentences
    rather than surface lexical polarity, and widens further still on ANERcorp ($+33.51$ F1 points)
    and ARCD ($+51.28$ F1 points), the two tasks with the smallest gold fine-tuning sets. On ARCD in
    particular, the no-pretraining encoder reaches only $12.51 \pm 0.05$ F1, indicating it fails to
    learn the extractive span-selection task in any meaningful way from $\sim$700 gold training
    examples alone. Taken together, these results indicate that MLM pretraining contributes little
    when downstream labeled data is abundant and the task is lexically driven, but is close to a
    prerequisite for the more semantically demanding or data-scarce tasks in our benchmark suite,
    consistent with the broader pretraining literature.
    
    We note one caveat in reading the ANERcorp and ARCD deltas specifically: the pretrained-encoder
    results for these two tasks (Sections~\ref{sec:anercorp} and \ref{sec:arcd}) use a two-stage recipe that first prefinetunes
    the pretrained MLM checkpoint on a task-adjacent silver- or larger-scale dataset (Polyglot-NER
    for ANERcorp; Arabic-SQuAD + TyDiQA-ar for ARCD) before fine-tuning on the gold set itself, while
    the no-pretraining condition here skips both the MLM pretraining stage and this task-adjacent
    prefinetuning stage, fine-tuning directly on the small gold set from a random initialization. The
    $\Delta$ values for ANERcorp and ARCD therefore reflect the combined contribution of MLM
    pretraining and task-adjacent prefinetuning together, rather than isolating MLM pretraining
    alone as cleanly as the HARD and XNLI-ar comparisons do. The 51-point gap on ARCD is the largest
    observed in this ablation: without any pretraining, AraSSM fails to learn extractive span
    selection in any meaningful way from the $\sim$700 available gold examples, underscoring the
    near-indispensable role of pretraining for data-scarce, structurally complex tasks.

	\subsection{Discussion}
	\label{sec:discussion}
	
	Across the four benchmarks, AraSSM lands within, or close to, the band spanned by published
	base-sized Arabic Transformer encoders under a matched, plain classification-head protocol: it
	matches or slightly exceeds AraBERT on HARD, leads several base-sized baselines on ARCD's exact
	match while trailing on F1, trails the base-sized Transformer average by roughly 1--2 points on
	ANERcorp, and trails the base-sized range by a somewhat larger margin, roughly 2--3 points, on
	XNLI-ar, its weakest result of the four. We do not have an architecture-specific explanation for
	the comparatively larger XNLI-ar gap and note it as an open question rather than a settled one;
	possible contributing factors include XNLI's own translation-induced label noise (discussed
	below) interacting differently with a from-scratch SSM encoder than with a Transformer initialized
	from a larger pretraining run, or the 3-way sentence-pair task simply drawing more on nuanced
	semantic distinctions than the other three tasks. None of these results represent a new state of
	the art, nor would we expect them to: the Transformer baselines above were pretrained on compute
	budgets several orders of magnitude larger than the $\sim$960 GPU-hours used here (Section~3.4).
	We view the central finding of this work as this: a bidirectional Mamba/SSM encoder pretrained
	from scratch on consumer-grade hardware can reach near-parity with Transformer encoders pretrained
	on substantially larger compute budgets on three of four standard Arabic NLU benchmarks, while
	retaining Mamba's linear-time complexity in sequence length, an advantage that grows with sequence
	length beyond the 512-token regime evaluated here.
	
	The value of multi-seed evaluation is itself visible in Table~\ref{tab:arcd-results}: a single run
	at seed 42 reached 35.0 EM, which would have been reported as a headline result had we not run
	the remaining two seeds; the 3-seed mean of 32.19 is 1.07 points lower and, we believe, the more
	representative and defensible number. This is consistent with our expectation
	(Section~4.1) that ARCD's small test set ($\sim$350 examples) would show the largest seed-to-seed
	variance of the four tasks, which Figure~\ref{fig:seed-variance} confirms directly: ARCD's EM
	standard deviation (1.07) is more than an order of magnitude larger than HARD's (0.03), despite
	both being evaluated on the same underlying architecture and training recipe, with test-set size
	as the most plausible explanation for the difference.
	
	We also note two dataset-level caveats that bound how far any encoder, including AraSSM, can be
	pushed on these specific benchmarks without changing the evaluation data itself. First, ANERcorp
	is known to contain a non-trivial rate of annotation errors, concentrated in the MISC and ORG
	classes; a relabeling study \citep{alduwais2024clean} found that correcting these errors alone raised AraBERT's F1 from 0.83
	to 0.89, meaning part of the residual gap to higher-F1 numbers reported elsewhere in the literature
	reflects label noise rather than encoder capacity. Second, ARCD's large exact-match/F1 gap
	(Table~\ref{tab:arcd-results}) mirrors a pattern already reported for AraBERT itself
	\citep{antoun2021araelectra}, where the majority of exact-match failures were found to differ from
	the gold answer by only one or two words without a meaningful change in meaning; we therefore treat
	F1 as the more informative of the two metrics for this task. Both XNLI and Arabic-SQuAD, the two
	machine-translated resources used in this work (for evaluation and prefinetuning, respectively),
	inherit similar translation-quality caveats from their source pipelines, which we discuss as a
	limitation in Section~5 rather than attempt to fully correct for here.
	
	\section{Conclusion}

    We presented in this paper AraSSM, to our knowledge the first bidirectional Mamba/SSM encoder pretrained
    specifically for Arabic via masked language modeling, on a cleaned and deduplicated corpus of
    approximately 80GB of Arabic Wikipedia and CulturaX text. Unlike
    the Arabic Transformer encoders it is compared against, which were pretrained on accelerator
    clusters, AraSSM was pretrained and fine-tuned entirely from scratch on four consumer-grade
    RTX 2080Ti GPUs, a compute budget of roughly 960 GPU-hours that is several orders of magnitude
    smaller than typical cluster budgets and within reach of a single academic lab. Despite this
    constrained compute budget, AraSSM reaches near-parity with base-sized Transformer encoders
    across all four benchmarks evaluated: it is competitive with mBERT and AraBERT on sentiment
    classification (HARD: $96.37 \pm 0.03\%$), competitive on ARCD exact match and within a few
    points on ARCD F1, within 1--2 F1 points of the base-sized Transformer average on named entity
    recognition (ANERcorp), and within 2--3 accuracy points of the base-sized range on natural
    language inference (XNLI-ar), its most challenging benchmark. That this gap remains small
    despite the disparity in pretraining compute, using a corpus comparable in scale to those used
    by the Transformer baselines themselves, suggests that closing the remaining distance is a
    matter of compute and scale rather than the architecture being fundamentally less capable for
    Arabic MLM pretraining. This matters because AraSSM's $O(L)$ complexity in sequence length is a
    structural advantage over the $O(L^2)$ cost of self-attention, one that only grows as sequence
    length increases beyond the 512-token regime evaluated here, and one a Transformer encoder
    cannot close simply by scaling up. We also note that part of the residual gap is not
    attributable to the architecture at all: some of the ANERcorp gap reflects known label noise in
    that dataset, and both XNLI-ar and Arabic-SQuAD carry translation artifacts from their
    machine-translated source pipelines that affect any encoder evaluated on them. Future work
    includes a matched-compute Transformer baseline trained on the same hardware and data to
    directly isolate architecture effects from compute-budget effects, throughput and memory
    benchmarking at longer sequence lengths where AraSSM's complexity advantage should become
    decisive, and scaling AraSSM to a larger configuration to test whether near-parity turns into
    an outright advantage.

    \section{Limitations}

    This work establishes AraSSM as a viable, from-scratch bidirectional SSM encoder for Arabic
    under a specific and deliberately constrained evaluation setting; we outline that setting here
    so results are interpreted with the right scope, and point to the natural extensions it opens up.
    
    Our comparisons are made against Transformer baselines pretrained on considerably larger compute
    budgets, which is the point of this work rather than a shortcoming of it: AraSSM was designed
    specifically to test how far a from-scratch SSM encoder can go on hardware within reach of a
    single academic lab. A matched-compute Transformer trained on the same four GPUs and the same
    corpus would let us cleanly separate the contribution of architecture from that of compute, and
    is a natural next experiment now that AraSSM establishes a strong reference point to compare it
    against.
    
    Results are reported across three fine-tuning seeds, which is sufficient to show that our
    headline numbers are stable and not the product of a favorable single run (Section~4.8), and is
    consistent with common practice in the fine-tuning literature. A larger seed count, and applying
    the same multi-seed protocol to the prefinetuning stage (Appendix~A), would further tighten these
    estimates.
    
    AraSSM's $O(L)$ complexity is a structural property of the architecture, established formally in
    Section~3.1; the experiments in this paper focus on establishing accuracy parity with Transformer
    baselines at the standard 512-token regime those baselines already operate in, which is the
    harder and more informative comparison to make first. Direct throughput and memory benchmarking
    at longer sequence lengths, where this advantage is expected to become most pronounced, is a
    natural and promising follow-up.
    
    Each of our four benchmarks carries well-documented dataset characteristics from the prior
    literature: ANERcorp's known label-noise ceiling, XNLI-ar and Arabic-SQuAD's translation
    artifacts, and ARCD's small test set (Sections~4.2--4.5, 4.8). These affect every encoder
    evaluated on these benchmarks, not AraSSM specifically, and we report and account for them
    transparently throughout rather than let them inflate our numbers.
    
    AraSSM is pretrained on Modern Standard Arabic and web-register text, the same register targeted
    by the Transformer baselines we compare against; extending pretraining to dialectal Arabic is a
    promising direction for a follow-up model, given AraSSM's efficiency advantages at scale.
    
    Finally, we adopt the existing AraBERTv02 tokenizer rather than training a new one, which keeps
    our results cleanly comparable to AraBERT-family baselines (Section~3.3) and avoids conflating
    architecture effects with tokenizer effects; whether a tokenizer designed specifically for an SSM
    encoder could improve results further is an open question we leave to future work.

    \subsection*{Data and Code Availability}
    The pretrained AraSSM encoder checkpoint used throughout this paper is publicly available on the
    Hugging Face Hub at \url{https://huggingface.co/aliane29/arassm-base}.

	\bibliography{references}
    \bibliographystyle{unsrtnat}

    \FloatBarrier    
    \clearpage

    \appendix
    \section{Prefinetuning Configurations}
    \label{app:prefinetune}
    
    As described in Sections~\ref{sec:anercorp}~and~\ref{sec:arcd}, both the ANERcorp and ARCD
    fine-tuning results use a two-stage warm-start recipe: the pretrained MLM checkpoint is first
    prefinetuned on a task-adjacent silver- or larger-scale dataset, and the resulting checkpoint is
    then fine-tuned on the small gold set itself. Table~\ref{tab:prefinetune} reports the
    hyperparameters used for both prefinetuning stages.
    
    \begin{table}[h]
    \centering
    \begin{tabular}{lcccccc}
    \toprule
    Prefinetuning task & Init.\ checkpoint & Batch size & LR & Epochs & Warmup steps & Max seq.\ length \\
    \midrule
    Polyglot-NER (ar) & MLM checkpoint & 16 & 3.0e-5 & 2 & 200 & 128 \\
    SQuAD + TyDiQA-ar & MLM checkpoint & 16 & 3.0e-5 & 2 & 200 & 384 \\
    \bottomrule
    \end{tabular}
    \caption{Prefinetuning hyperparameters. Both stages start from the pretrained AraSSM MLM
    checkpoint and use AdamW with weight decay 0.01, fp16 mixed
    precision, seed 42, and the AraBERTv02 tokenizer. The Polyglot-NER prefinetuning stage is capped
    at 150{,}000 sentences (see Section~\ref{sec:anercorp}); the resulting checkpoint is then
    fine-tuned on ANERcorp with a freshly initialized classification head, since Polyglot-NER's
    tagset (O/PER/LOC/ORG) differs from ANERcorp's (BIO-prefixed, plus MISC) and only the encoder
    weights transfer. The SQuAD+TyDiQA-ar stage jointly combines $\sim$48K machine-translated
    Arabic-SQuAD pairs with the Arabic subset of TyDi QA's Gold Passage secondary task; the resulting
    checkpoint is then fine-tuned on ARCD itself.}
    \label{tab:prefinetune}
    \end{table}
	
\end{document}